\documentclass[conference]{IEEEtran}
\IEEEoverridecommandlockouts
\usepackage{cite}
\usepackage{amsmath,amssymb,amsfonts}
\usepackage{placeins}
\usepackage{algorithmic}
\usepackage{graphicx}
\usepackage{textcomp}
\usepackage{xcolor}
\usepackage{float}
\usepackage{booktabs,multirow}
\def\BibTeX{{\rm B\kern-.05em{\sc i\kern-.025em b}\kern-.08em
    T\kern-.1667em\lower.7ex\hbox{E}\kern-.125emX}}
\begin{document}

\title{Preventing Overfitting in Deep Image Prior for Hyperspectral Image Denoising\\
\thanks{This research work was supported by the project “Applied Research for Autonomous Robotic Systems” (MIS 5200632) which is implemented within the framework of the National Recovery and Resilience Plan “Greece 2.0” (Measure: 16618- Basic and Applied Research) and is funded by the European Union- NextGenerationEU.}
}

\author{
\IEEEauthorblockN{
Panagiotis~Gkotsis\IEEEauthorrefmark{1} and 
Athanasios A. Rontogiannis\IEEEauthorrefmark{1}\IEEEauthorrefmark{2}
}
\IEEEauthorblockA{
\IEEEauthorrefmark{1}Robotics Institute, Athena Research and Innovation Center, 151~25 Maroussi, Greece\\
\IEEEauthorrefmark{2}School of Electrical and Computer Engineering, National Technical University of Athens, 157~80 Athens, Greece\\
Emails: p.gkotsis@athenarc.gr, th\_rontogiannis@mail.ntua.gr
}
}


\maketitle

\begin{abstract}
Deep image prior (DIP) is an unsupervised deep learning framework that has been successfully applied to a variety of inverse imaging problems. However, DIP-based methods are inherently prone to overfitting, which leads to performance degradation and necessitates early stopping. In this paper, we propose a method to mitigate overfitting in DIP-based hyperspectral image (HSI) denoising by jointly combining robust data fidelity and explicit sensitivity regularization. The proposed approach employs a Smooth $\ell_1$ data term together with a divergence-based regularization and input optimization during training. Experimental results on real HSIs corrupted by Gaussian, sparse, and stripe noise demonstrate that the proposed method effectively prevents overfitting and achieves superior denoising performance compared to state-of-the-art DIP-based HSI denoising methods.
\end{abstract}


\section{Introduction}

Hyperspectral imaging captures scene information across multiple spectral channels, typically spanning visible to infrared wavelengths. It has become a powerful technology with applications in environmental monitoring, precision agriculture, planetary exploration, food quality assessment, and medicine, among others \cite{ghamisi2017, bhargava2024}. In practice, hyperspectral images (HSIs) are often degraded by sensor limitations and atmospheric effects, which introduce noise such as Gaussian, sparse, or striping artifacts. Effective denoising is therefore essential and commonly employed as a critical pre-processing step in hyperspectral imaging applications.

A thorough overview of modern HSI denoising techniques is presented in several recent reviews, e.g. \cite{zzys23}, \cite{joglekar2025}.  Such techniques can generally be classified into model-based and deep learning (DL) methods. Model-driven approaches usually formulate denoising as an optimization problem and leverage prior assumptions about the spatial-spectral structure of HSIs, such as low-rankness and sparsity. This is often achieved by incorporating suitable regularization terms in the cost function, e.g., $\ell_1$ norm and total variation (TV) \cite{fan2018}. In contrast, DL-based methods are data-driven, that is, they typically utilize 2D or 3D convolutional neural network (CNN) architectures to learn the complex spatial-spectral noise patterns directly from data, without explicitly defining hand-crafted priors \cite{yuan2019}. This approach often achieves improved results over model-based methods. However, in practice, such supervised learning models might not be as successful in real-world scenarios, due to the lack of large training datasets containing denoised target images, as well as the variety of noise conditions in real HSIs, which can hamper the model's ability to generalize.

Unsupervised approaches based on the deep image prior (DIP) framework \cite{ulyanov2018} offer an appealing alternative. DIP exploits the implicit bias of convolutional neural network architectures, using a randomly initialized network trained on a single corrupted image. This idea has been extended to hyperspectral data in the deep hyperspectral image prior (DHIP) model \cite{sidorov2019}. Despite their effectiveness, DIP-based methods are inherently prone to overfitting, i.e.,  as training progresses, the network eventually memorizes noise, leading to performance degradation and necessitating careful early stopping.

Several strategies have been proposed to mitigate overfitting in DIP. Some approaches introduce architectural constraints or additional regularization terms \cite{tirer2024,alkhouri2025,liang2025,li2023}, while others incorporate Stein’s unbiased risk estimator (SURE) \cite{stein1981} into the loss function to penalize sensitivity to noise through a divergence term \cite{metzler2020,nguyen2021}. SURE-based methods effectively suppress overfitting under Gaussian noise but rely on quadratic data fidelity and are less robust to mixed or heavy-tailed noise commonly encountered in HSIs. Conversely, robust $\ell_1$-type data fidelity terms have been shown to significantly improve denoising performance under sparse and non-Gaussian noise \cite{niresi2022}, yet when used alone in DIP training they do not prevent overfitting and still require early stopping.

It should be emphasized that robust data fidelity and divergence-based regularization are not trivially compatible in the DIP setting. Robust losses alter the geometry of the optimization landscape and may even accelerate memorization in highly overparameterized networks, while divergence regularization has primarily been studied under $\ell_2$-based formulations. Moreover, recent works have shown that jointly optimizing the network input can improve convergence and reconstruction quality \cite{liang2025,li2023}, but this substantially increases the risk of overfitting, as noise can be encoded both in the network parameters and in the learned input representation.

In this work, we show that effective overfitting mitigation in DIP-based HSI denoising requires the joint design of representation learning and sensitivity regularization. We propose a unified loss formulation that combines a Smooth $\ell_1$ data fidelity term with a SURE-inspired divergence regularization, and we employ joint optimization over the network parameters and the input image. The Smooth $\ell_1$ term provides robustness to mixed and structured noise, while the divergence term explicitly penalizes sensitivity to perturbations, acting as a regularizer on the estimator’s degrees of freedom. Crucially, we demonstrate that divergence regularization becomes effective only in conjunction with input optimization, where it stabilizes the expanded optimization space and prevents noise memorization.
Experimental results on real hyperspectral data under Gaussian, sparse, and striping noise confirm that the proposed method consistently prevents overfitting and achieves superior denoising performance compared to state-of-the-art DIP-based approaches, without reliance on early stopping. 

\section{Deep Hyperspectral Image Prior}

A hyperspectral image is a 3-D array (tensor) of dimensions $w \times h \times b$, where $w$, $h$ are the width and height of the image and $b$ is the number of spectral bands. We now consider a noisy vectorized HSI $\mathbf{y} \in \mathbb{R}^n$, where $n = w\cdot h\cdot b$,  and denote its clean counterpart as $\mathbf{x} \in \mathbb{R}^n$. In the denoising problem, the noisy and clean images are assumed to be related by the observation model
\begin{equation}
    \mathbf{y} = \mathbf{x} + \mathbf{n},
\end{equation}
where $\mathbf{n}$ is additive noise. In general, $\mathbf{n}$ can consist of more than one different noise types, e.g.,
\begin{equation}
    \mathbf{n} = \mathbf{w} + \mathbf{s},
\end{equation}
where $\mathbf{w} \sim \mathcal{N}(0, \sigma^2 \mathbf{I}_n)$ is i.i.d. Gaussian noise and $\mathbf{s}$ corresponds to sparse noise. HSI denoising seeks to reconstruct a clean image $\mathbf{x}$ from the noisy  observation $\mathbf{y}$, such that the reconstructed image $\mathbf{\hat{x}}$  closely approximates $\mathbf{x}$. 

In the deep hyperspectral image prior (DHIP) method introduced in \cite{sidorov2019}, which is based directly on the original DIP approach \cite{ulyanov2018}, the HSI denoising problem is tackled with a U-Net-type CNN architecture, i.e., encoder-decoder, with skip connections. The input to the network is an HSI $\mathbf{z}$ drawn from a Gaussian distribution (or the noisy observation $\mathbf{y}$ as in \cite{metzler2020}) and the output is referred to as $f_{\boldsymbol{\theta}}(\mathbf{z})$. The estimate of the original image is then given by
\begin{equation}
    \mathbf{\hat{x}} = f_{\boldsymbol{\hat{\theta}}}(\mathbf{z}),
\end{equation}
where $\boldsymbol{\hat{\theta}}$ is the network parameter vector which minimizes the loss function:
\begin{equation}
    \boldsymbol{\hat{\theta}} = \arg\,\min_{\boldsymbol{\theta}} \mathcal{L}(\boldsymbol{\theta}; \mathbf{y}).
\end{equation}
The loss function employed in the original DIP and DHIP models is the $\ell_2$ loss
\begin{equation}
    \mathcal{L}(\boldsymbol{\theta}; \mathbf{y}) = \lVert f_{\boldsymbol{\theta}}(\mathbf{z}) - \mathbf{y} \rVert_2^2.
\end{equation}

The effectiveness of the DIP arises from its convolutional architecture, which acts as an implicit prior for the structure of natural images. Specifically, it inherently favors the representation of structured patterns such as those found in images and, as a result, is more efficient at capturing the clean image than the imposed noise. Consequently, as the training progresses, the network produces a denoised image before it begins fitting the noise. This behavior renders DIP suitable for denoising, but at the same time gives rise to overfitting, causing degradation of the output image due to the fitting of noise. This imposes a limit on the number of iterations that one should allow the model to run, which is neither known nor consistent between different denoising tasks. This issue is the main limitation of the method.

\begin{figure}[t]
    \centering
    \includegraphics[width=\linewidth]{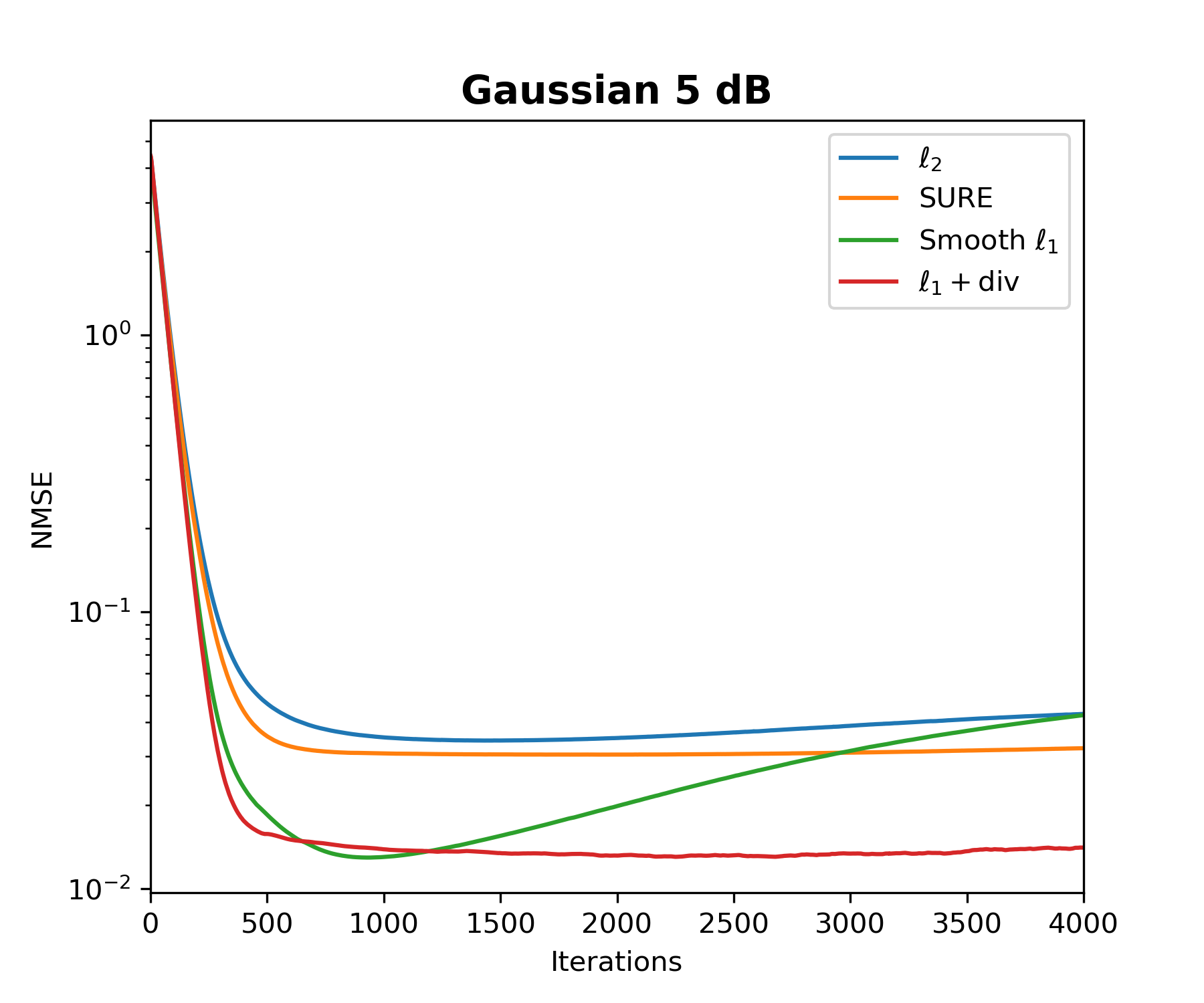}
    \caption{NMSE for a training period of 4000 iterations using the DHIP model architecture \cite{sidorov2019} with different loss functions.}
    \label{fig:1}
\end{figure}

The overfitting behavior of the model is illustrated in Fig. \ref{fig:1}, where the normalized mean square error (NMSE) is showcased for trainings of the DHIP model architecture with different loss functions. In this experiment, $\mathbf{y}$ is the Washington DC mall (DC) HSI dataset, with additive Gaussian noise of $\text{SNR} = 5 \text{dB}$. The training curve of DHIP with the $\ell_2$ loss (in blue) can be seen to exhibit overfitting.

\begin{figure*}[t]\footnotesize
    \begin{minipage}[b]{0.32\linewidth}
        \centering
        \centerline{\includegraphics[width=6.0cm]{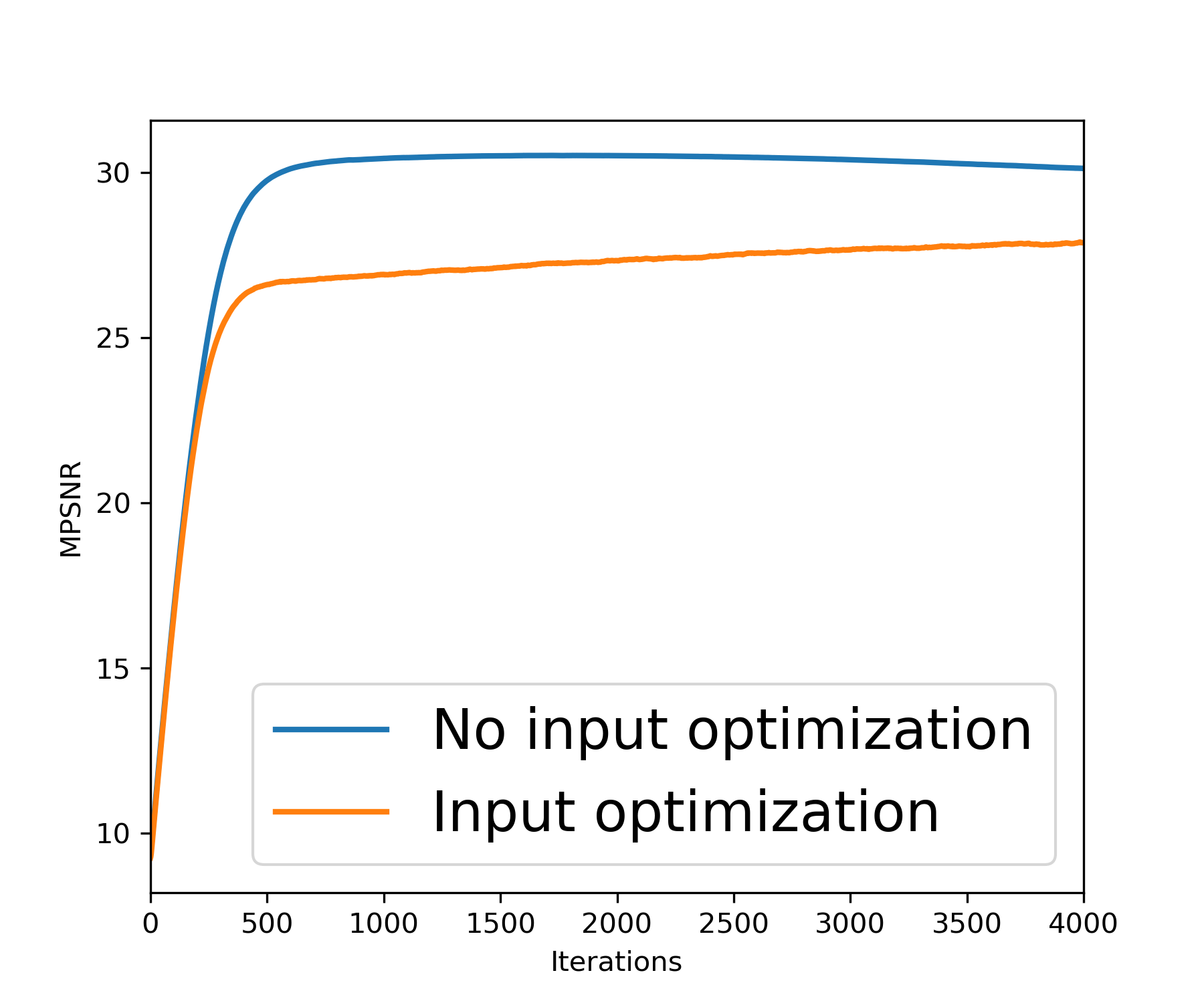}}
        \centerline{(a) SURE}
    \end{minipage}
    \hfill
    \begin{minipage}[b]{0.32\linewidth}
        \centering
        \centerline{\includegraphics[width=6.0cm]{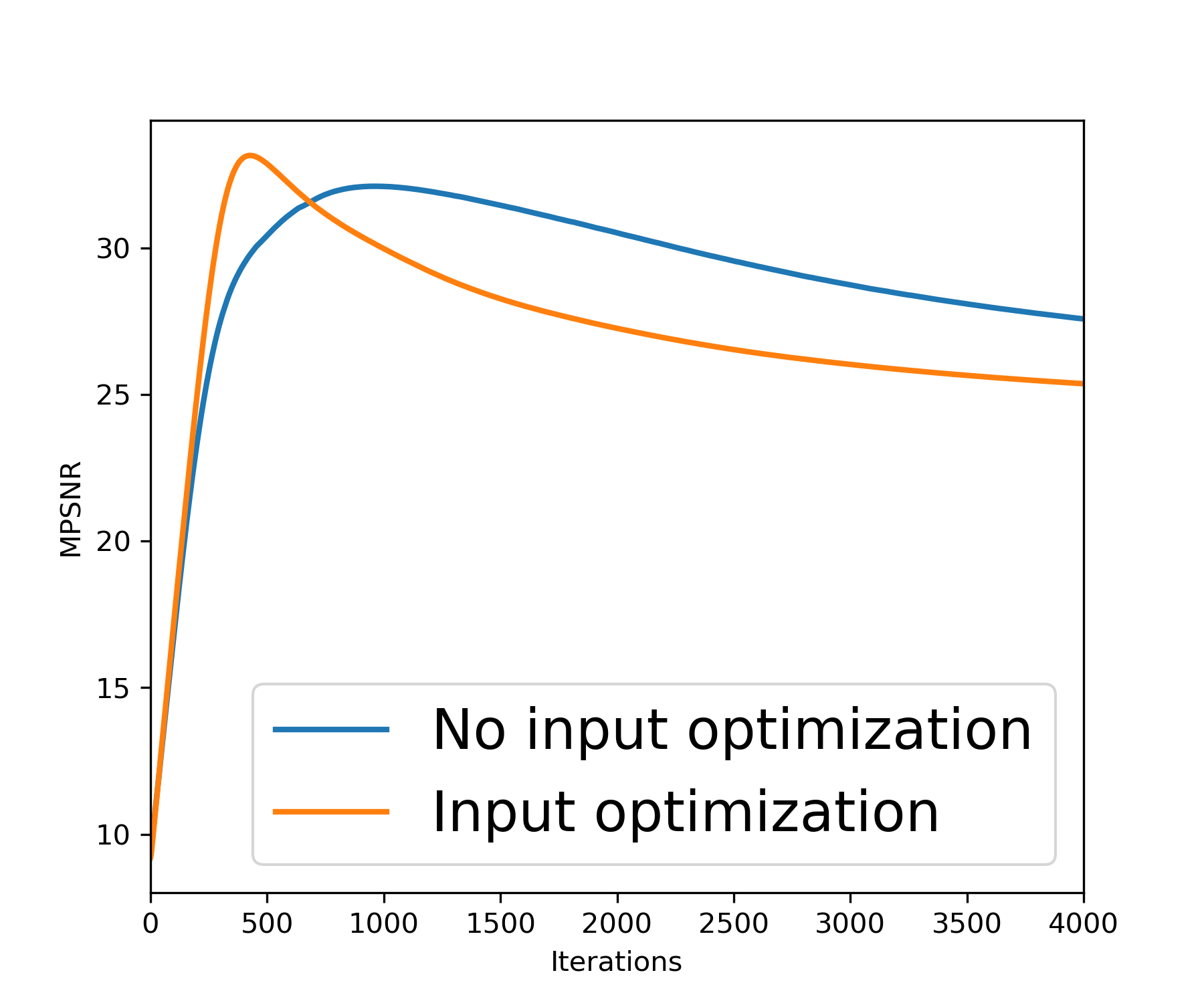}}
        \centerline{(b) Smooth-$\ell_1$}
    \end{minipage}
    \hfill
    \begin{minipage}[b]{0.32\linewidth}
        \centering
        \centerline{\includegraphics[width=6.0cm]{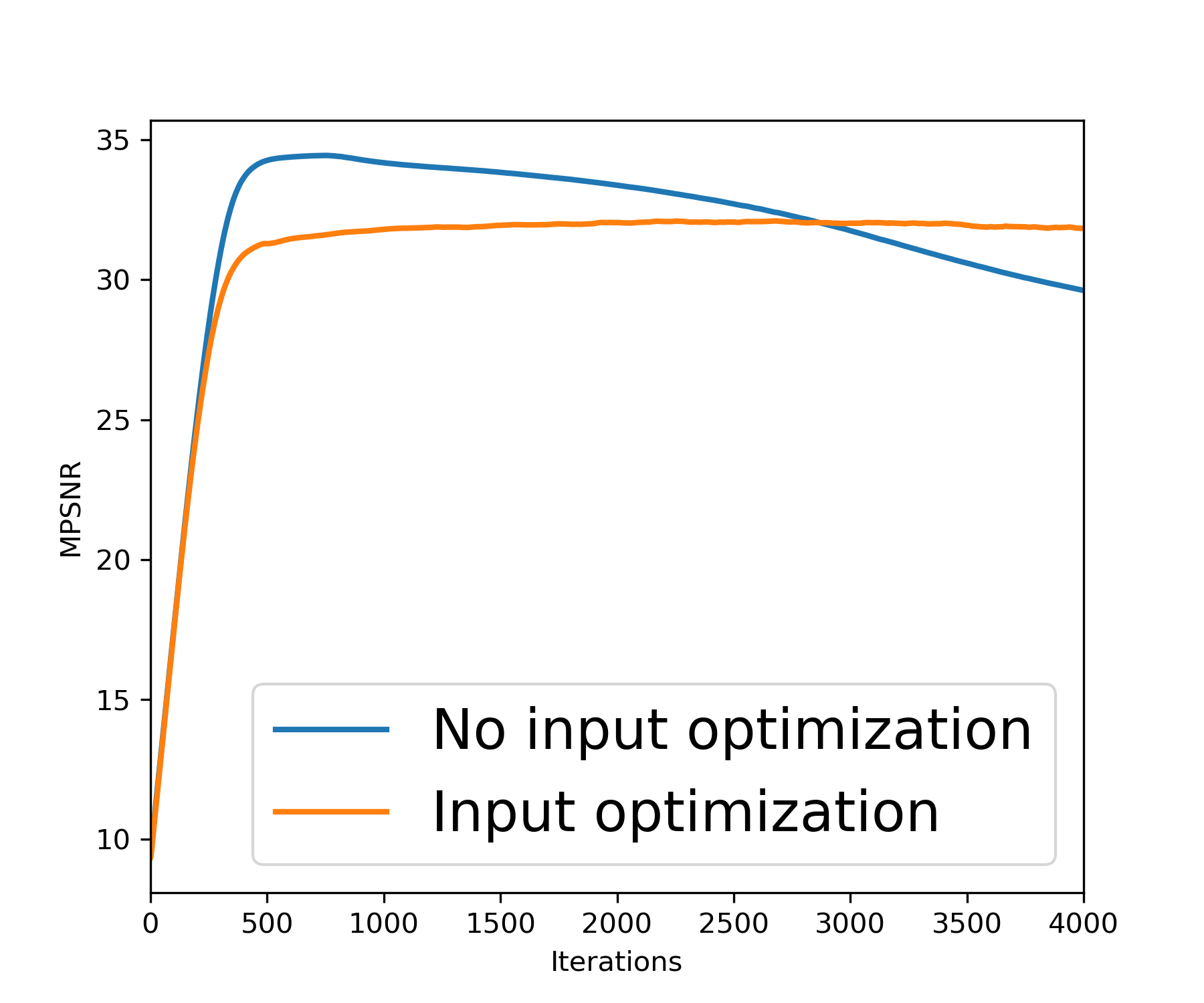}}
        \centerline{(c) $\ell_1 + \text{div}$}
    \end{minipage}
    \caption{Effect of joint input optimization on denoising performance under additive Gaussian noise for three different loss formulations.}
    \label{fig:2}
\end{figure*}
\section{Preventing Overfitting in DHIP}
As discussed in Section II, the DHIP framework is inherently prone to overfitting, as the overparameterized network progressively memorizes noise during optimization. In this section, we describe a strategy to mitigate this behavior by explicitly controlling estimator sensitivity while preserving robustness to mixed and structured noise. The proposed approach combines divergence-based sensitivity regularization, robust data fidelity, and joint optimization of the network parameters and the input representation.

\subsection{Divergence-Based Sensitivity Regularization}

Stein’s unbiased risk estimator (SURE) \cite{stein1981} provides an estimate of the mean squared error (MSE) using only the noisy observation $\mathbf{y}$, under additive Gaussian noise assumptions. For an estimator $f_{\boldsymbol{\theta}}(\cdot)$ the SURE objective is given by
\begin{equation}
    \mathcal{L}_{\text{SURE}}(\boldsymbol{\theta}) = \frac{1}{n} \lVert \mathbf{y} - f_{\boldsymbol{\theta}}(\mathbf{y}) \rVert_2^2 - \sigma^2 + \frac{2\sigma^2}{n} \text{div}_{\mathbf{y}}(f_{\boldsymbol{\theta}}(\mathbf{y})),
    \label{eq:sure}
\end{equation}
where $\text{div}_{\mathbf{y}}(f_{\boldsymbol{\theta}}(\mathbf{y}))$ is the divergence term defined as
\begin{equation}
    \text{div}_{\mathbf{y}}(f_{\boldsymbol{\theta}}(\mathbf{y})) = \sum_{i=1}^n \frac{\partial f_{\theta i}(\mathbf{y})}{\partial y_i}.
\end{equation}
The divergence term penalizes estimators that are sensitive to perturbations of the noisy input and can be interpreted as a regularization on the estimator’s effective degrees of freedom \cite{stein1981}. In DIP-based methods with input equal to the noisy observation $\mathbf{y}$, this term discourages the network from fitting fine-scale perturbations in $\mathbf{y}$, thereby mitigating overfitting.

Since direct computation of the divergence is infeasible for deep neural networks, we employ the Monte Carlo approximation \cite{ramani2008}
\begin{equation}
    \text{div}_{\mathbf{y}}(f_{\boldsymbol{\theta}}(\mathbf{y})) = \mathbf{b}^T \left( \frac{f_{\boldsymbol{\theta}}(\mathbf{y} + \epsilon \mathbf{b}) - f_{\boldsymbol{\theta}}(\mathbf{y})}{\epsilon} \right),
    \label{eq:8}
\end{equation}
where $\mathbf{b} \sim \mathcal{N}(0, \mathbf{I}_n)$ is an i.i.d. Gaussian random vector and $\epsilon$ is a small number set to $10^{-3}$. As illustrated in Fig.~\ref{fig:1}, incorporating divergence regularization within a quadratic data-fidelity framework effectively suppresses overfitting under Gaussian noise conditions. However, this formulation relies on $\ell_2$-based data fidelity and is therefore sensitive to non-Gaussian, sparse, or structured noise.

\subsection{Robust Data Fidelity via Smooth $\ell_1$ Loss}
To improve robustness under mixed noise, we adopt the Smooth $\ell_1$ loss (also known as the Moreau envelope of the $\ell_1$ norm \cite{Theo25}), defined element-wise as
\begin{equation}
\mathcal{L}_{\text{Smooth-}\ell_1}(x, y; \beta) =
    \begin{cases}
        \frac{1}{2\beta}(x - y)^2, &\quad \lvert x - y \rvert \leq \beta\\
        \lvert x - y \rvert - \frac{\beta}{2}, &\quad \lvert x - y \rvert > \beta
    \end{cases}
    \label{eq:10}
\end{equation}
where $\beta$ (set to $10^{-3}$ in our experiments) is the threshold at which the loss changes between $\ell_1$ and $\ell_2$. 

The Smooth $\ell_1$ loss improves robustness to sparse and heavy-tailed noise while maintaining stable optimization dynamics. As shown by the green curve in Fig.~\ref{fig:1}, replacing the quadratic data-fidelity term with the Smooth $\ell_1$ loss significantly improves denoising performance. Nevertheless, robust data fidelity alone does not prevent overfitting, since despite its robustness to outliers, the estimator can still memorize noise during training, as the loss does not explicitly constrain estimator sensitivity. This observation highlights that robustness in the data-fidelity term and robustness to overfitting are fundamentally distinct properties.

\subsection{Unified Loss with Joint Input Optimization}
Motivated by the above observations, we propose a unified loss formulation that combines robust data fidelity with explicit sensitivity regularization under joint optimization of the network parameters and the input, i.e.,  
\begin{equation}
\mathcal{L}(\boldsymbol{\theta}, \mathbf{z})
=
\mathcal{L}_{\text{Smooth-}\ell_1}\!\big(f_{\boldsymbol{\theta}}(\mathbf{z}),\, \mathbf{y}\big)
+
\frac{2\sigma^2}{n} \operatorname{div}_{\mathbf{z}}\!\big(f_{\boldsymbol{\theta}}(\mathbf{z})\big).
\label{eq:uloss}
\end{equation}
Here,  
$\mathbf{z}$ denotes the optimized network input and the divergence term is penalized with the noise variance $\sigma^2$ as in (\ref{eq:sure}). The parameter $\sigma$ can be computed as described in \cite{nguyen2021}, by calculating the band-wise estimate
 \begin{equation}
     \hat{\sigma}_i = \frac{\text{median}\left(\lvert \mathbf{W}_{(i)}^{HH} \rvert\right)}{0.6745}, \quad i = 1, \dots, b
 \end{equation}
 using the median absolute deviation estimator in the highest subband (HH) of the wavelet transform of each band  and then taking the mean across the bands \cite{mallat2008}.

In the proposed loss function, the divergence term should not be interpreted as providing an unbiased estimate of the reconstruction risk. Instead, it acts as a sensitivity regularizer that constrains the local response of the estimator to perturbations of its current input representation. This role becomes particularly important under joint input optimization, where overfitting can otherwise arise through both the network parameters and the learned input.
As demonstrated by the red curve in Fig.~\ref{fig:1} and the orange curves in Fig.~\ref{fig:2}, which together form an ablation across loss formulations and input optimization, only the proposed combination of Smooth $\ell_1$ data fidelity, divergence-based sensitivity regularization, and joint input optimization achieves high denoising performance, stable convergence, and (as verified in Section IV) robustness across different noise scenarios.

\section{Experimental Results}

In this section, we evaluate the proposed method on hyperspectral image denoising under diverse noise conditions. All experiments are conducted on a $200\times200\times191$ segment of the Washington DC Mall and a $200\times200\times204$ segment of the Salinas HSI, corrupted according to the data model $\mathbf{y}=\mathbf{x}+\mathbf{n}$, where $\mathbf{n}$ may include Gaussian, sparse, and/or stripe noise. The proposed approach is compared with two DHIP-based denoising methods, namely SURE-DHIP \cite{nguyen2021}, which leverages Stein’s Unbiased Risk Estimator to mitigate overfitting, and HLF-DHIP \cite{niresi2022}, which employs a Smooth-$\ell_1$ loss to better handle diverse noise types. Performance is evaluated using the mean peak signal-to-noise ratio (MPSNR) and the mean structural similarity index (MSSIM).

\begin{figure}[t]
    \begin{minipage}[b]{0.49\linewidth}
        \centering
        \centerline{\includegraphics[width=4.5cm]{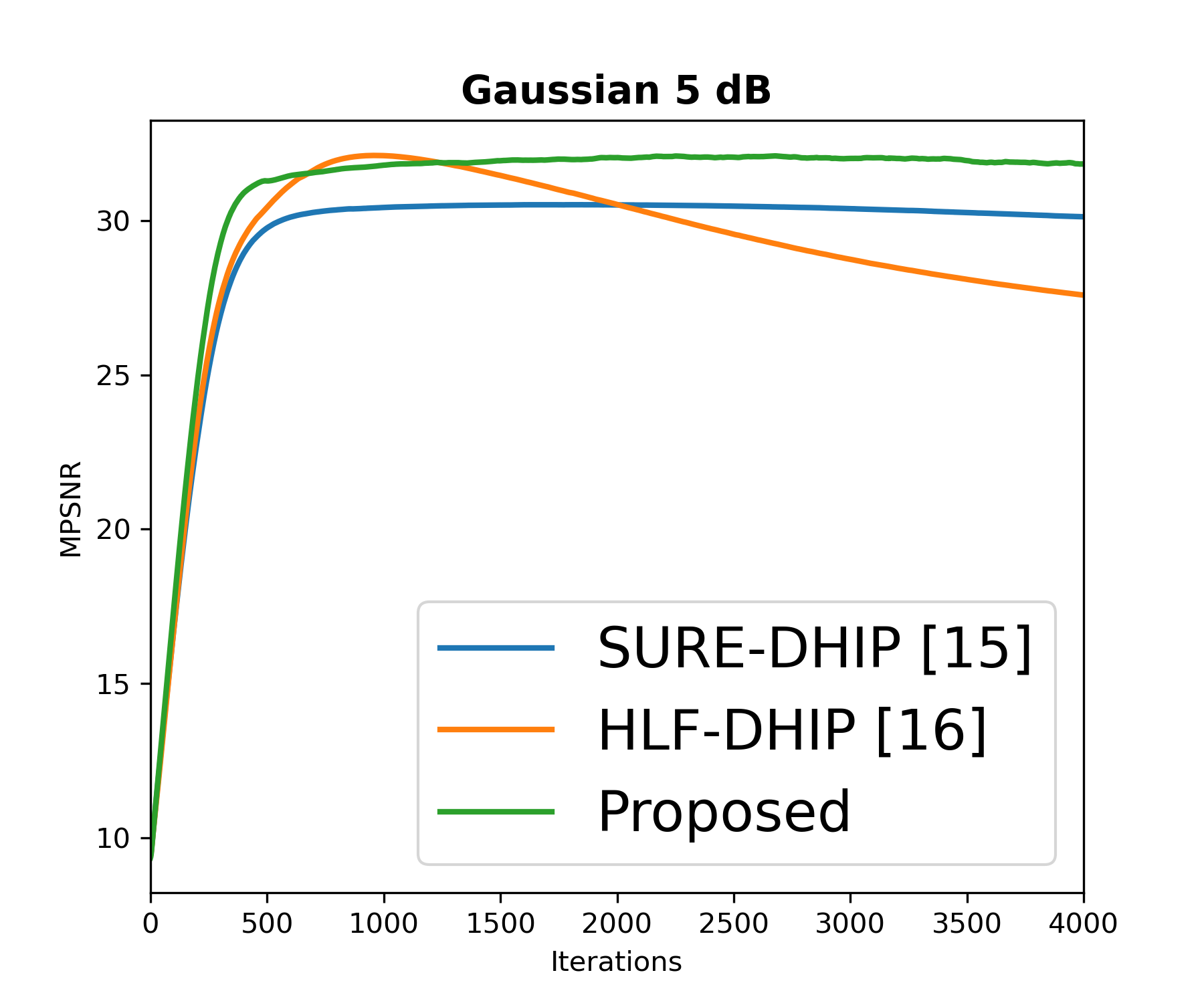}}
    \end{minipage}
    \hfill
    \begin{minipage}[b]{0.49\linewidth}
        \centering
        \centerline{\includegraphics[width=4.5cm]{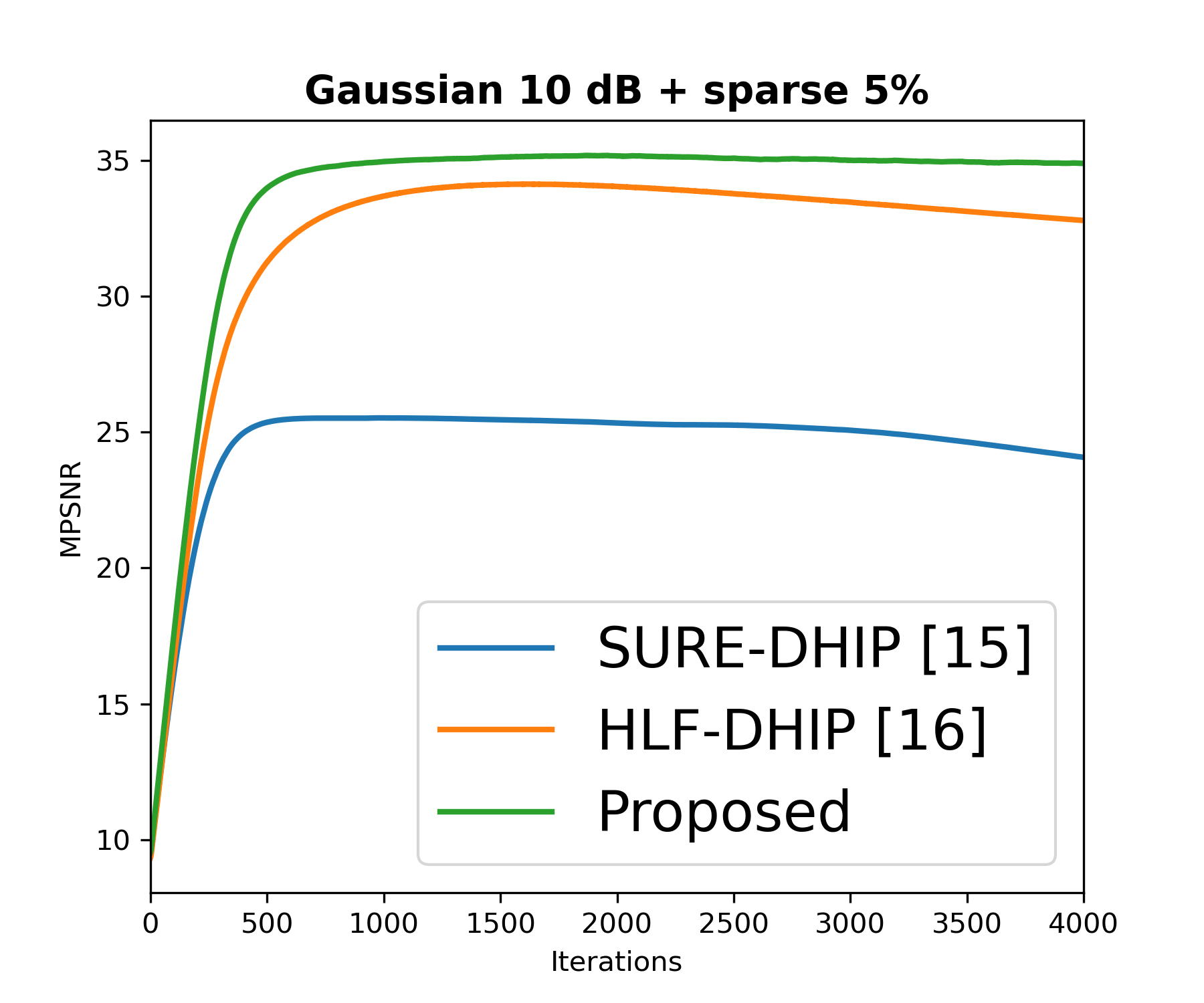}}
    \end{minipage}
    \\
    \begin{minipage}[b]{0.49\linewidth}
        \centering
        \centerline{\includegraphics[width=4.5cm]{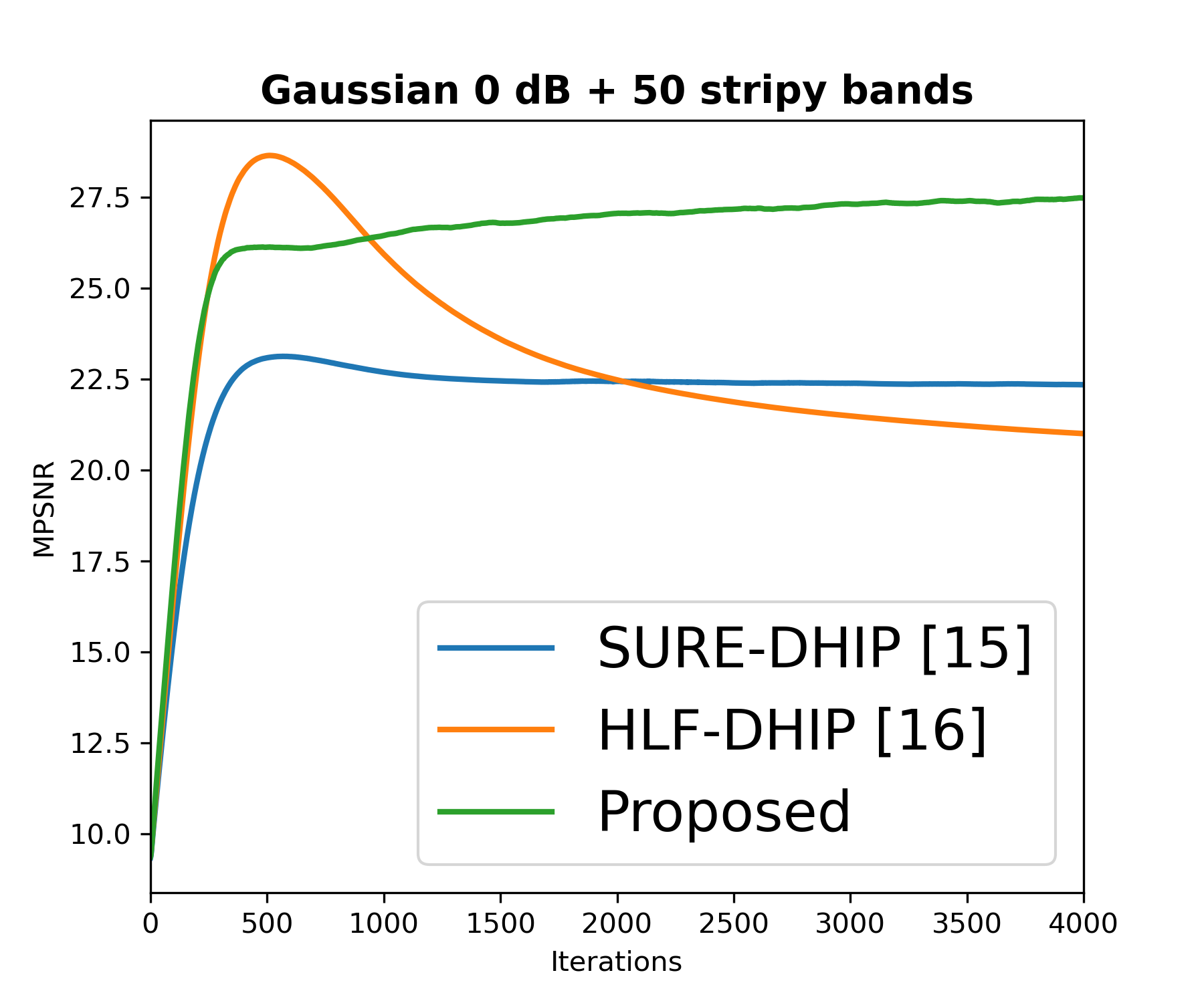}}
    \end{minipage}
    \hfill
    \begin{minipage}[b]{0.49\linewidth}
        \centering
        \centerline{\includegraphics[width=4.5cm]{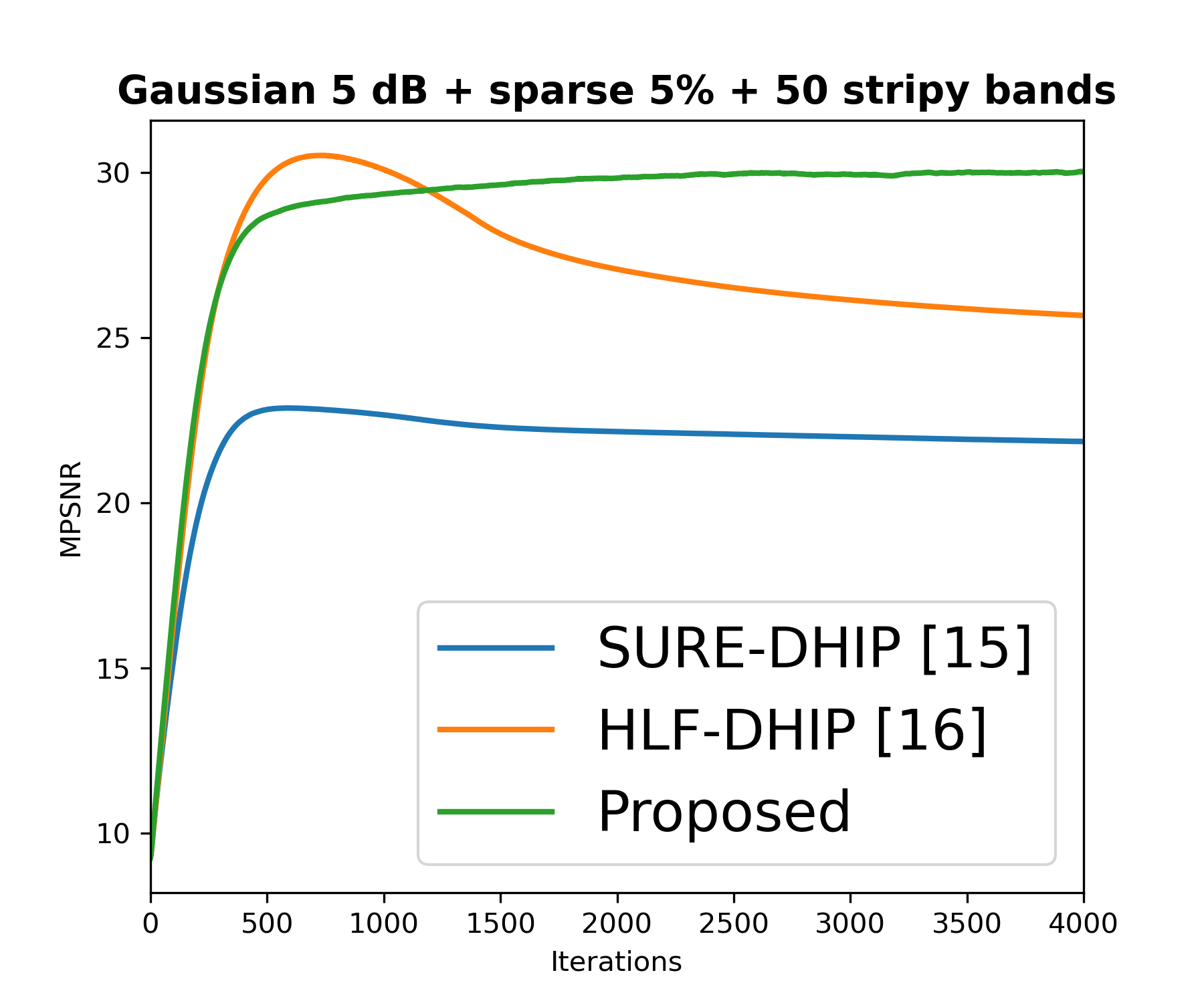}}
    \end{minipage}
    \caption{MPSNR results for SURE-DHIP \cite{nguyen2021}, HLF-DHIP \cite{niresi2022} and the proposed algorithms for various noise scenarios.}
    \label{fig:3}
\end{figure}

All methods are trained for 4000 iterations to allow both optimal reconstruction and potential overfitting effects to manifest. Figure \ref{fig:3} reports the MPSNR evolution on the DC Mall HSI for four noise scenarios: (i) Gaussian noise with SNR = 5 dB, (ii) Gaussian noise with SNR = 10 dB combined with sparse noise affecting 5\% of the pixels, (iii) Gaussian noise with SNR = 0 dB combined with stripe noise affecting 50 randomly selected spectral bands, and (iv) a mixture of Gaussian, sparse, and stripe noise. While HLF-DHIP significantly outperforms SURE-DHIP, particularly in the presence of sparse noise, it still exhibits clear overfitting behavior across all scenarios. In contrast, the proposed method consistently achieves higher peak performance and maintains stable convergence without degradation.

\begin{figure*}[ht]\footnotesize
    \begin{minipage}[b]{0.19\linewidth}
        \centering
        \centerline{\includegraphics[width=3.3cm]{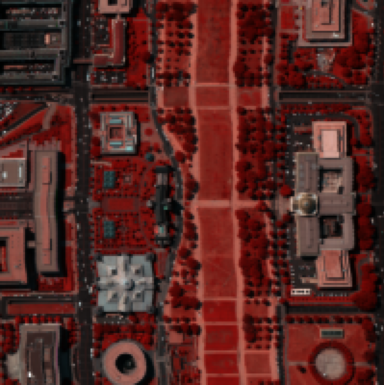}}
        \medskip
    \end{minipage}
    \hfill
    \begin{minipage}[b]{0.19\linewidth}
        \centering
        \centerline{\includegraphics[width=3.3cm]{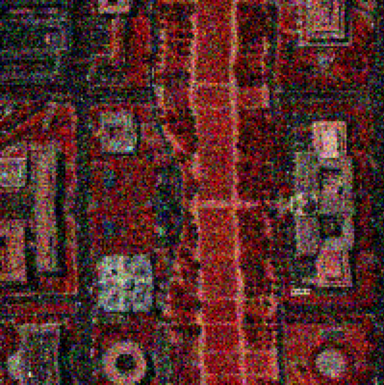}}
        \medskip
    \end{minipage}
    \hfill
    \begin{minipage}[b]{0.19\linewidth}
        \centering
        \centerline{\includegraphics[width=3.3cm]{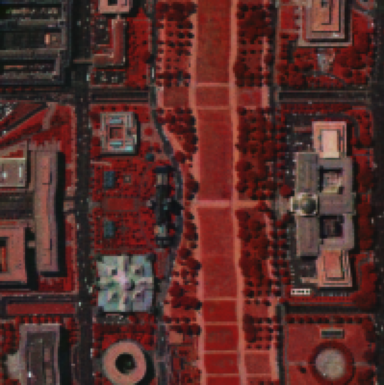}}
        \medskip
    \end{minipage}
    \hfill
    \begin{minipage}[b]{0.19\linewidth}
        \centering
        \centerline{\includegraphics[width=3.3cm]{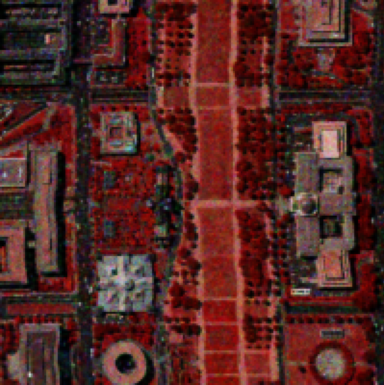}}
        \medskip
    \end{minipage}
    \hfill
    \begin{minipage}[b]{0.19\linewidth}
        \centering
        \centerline{\includegraphics[width=3.3cm]{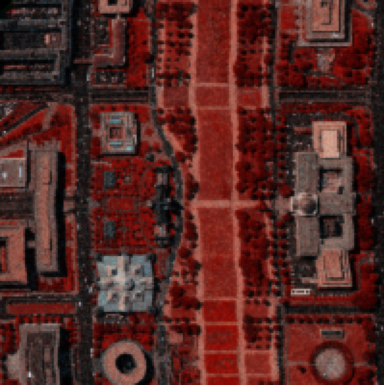}}
        \medskip
    \end{minipage}
    \\
    \begin{minipage}[b]{0.19\linewidth}
        \centering
        \centerline{\includegraphics[width=3.3cm]{media/4-1.png}}
        \medskip
    \end{minipage}
    \hfill
    \begin{minipage}[b]{0.19\linewidth}
        \centering
        \centerline{\includegraphics[width=3.3cm]{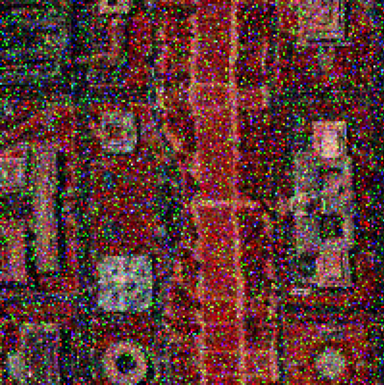}}
        \medskip
    \end{minipage}
    \hfill
    \begin{minipage}[b]{0.19\linewidth}
        \centering
        \centerline{\includegraphics[width=3.3cm]{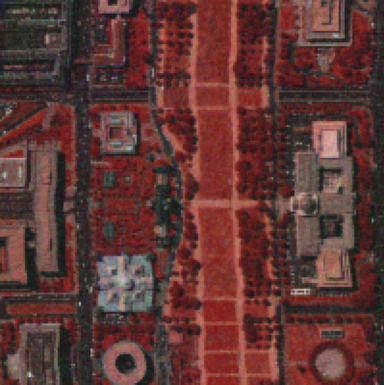}}
        \medskip
    \end{minipage}
    \hfill
    \begin{minipage}[b]{0.19\linewidth}
        \centering
        \centerline{\includegraphics[width=3.3cm]{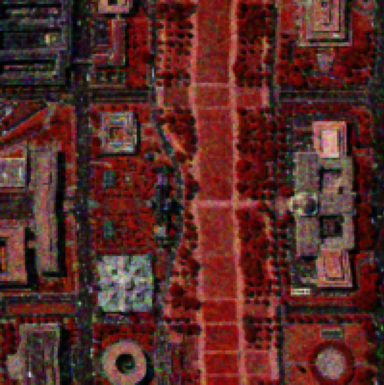}}
        \medskip
    \end{minipage}
    \hfill
    \begin{minipage}[b]{0.19\linewidth}
        \centering
        \centerline{\includegraphics[width=3.3cm]{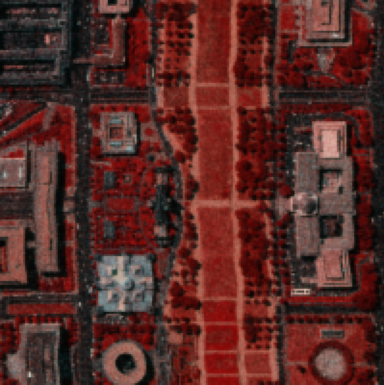}}
        \medskip
    \end{minipage}
    \\
    \begin{minipage}[b]{0.19\linewidth}
        \centering
        \centerline{\includegraphics[width=3.3cm]{media/4-1.png}}
        \centerline{(a) Original}
    \end{minipage}
    \hfill
    \begin{minipage}[b]{0.19\linewidth}
        \centering
        \centerline{\includegraphics[width=3.3cm]{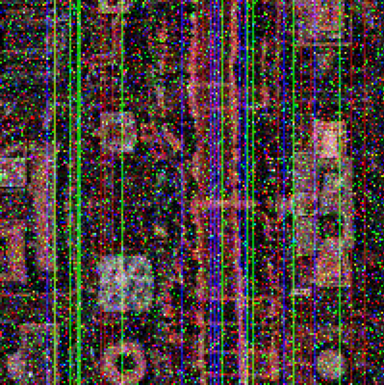}}
        \centerline{(b) Noisy}
    \end{minipage}
    \hfill
    \begin{minipage}[b]{0.19\linewidth}
        \centering
        \centerline{\includegraphics[width=3.3cm]{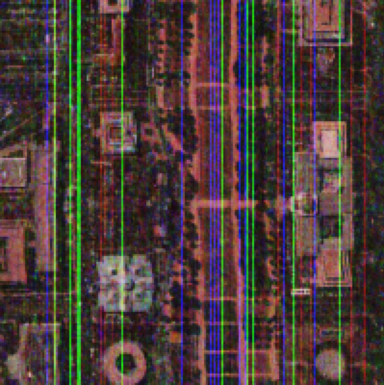}}
        \centerline{(c) SURE-DHIP \cite{nguyen2021}}
    \end{minipage}
    \hfill
    \begin{minipage}[b]{0.19\linewidth}
        \centering
        \centerline{\includegraphics[width=3.3cm]{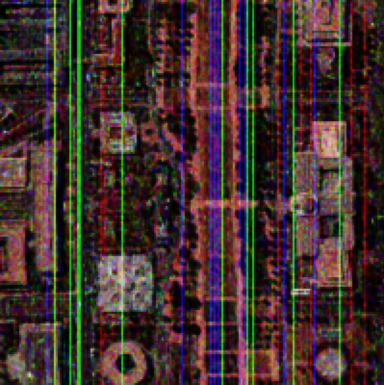}}
        \centerline{(d) HLF-DHIP \cite{niresi2022}}
    \end{minipage}
    \hfill
    \begin{minipage}[b]{0.19\linewidth}
        \centering
        \centerline{\includegraphics[width=3.3cm]{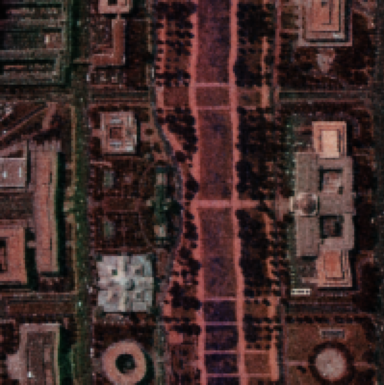}}
        \centerline{(e) Proposed}
    \end{minipage}
    \caption{Original Washington DC Mall HSI segment, noisy version, and denoised versions obtained by SURE-DHIP \cite{nguyen2021}, HLF-DHIP \cite{niresi2022} and the proposed method. The noise content in each row is Gaussian, Gaussian + sparse, and Gaussian + sparse + stripes for rows 1, 2, and 3 respectively.}
    \label{fig:4}
\end{figure*}

\begin{figure*}[ht]\footnotesize
    \begin{minipage}[b]{0.19\linewidth}
        \centering
        \centerline{\includegraphics[width=3.3cm]{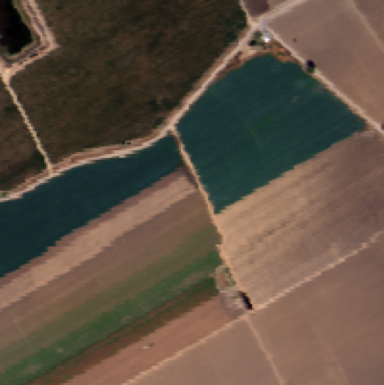}}
        \medskip
    \end{minipage}
    \hfill
    \begin{minipage}[b]{0.19\linewidth}
        \centering
        \centerline{\includegraphics[width=3.3cm]{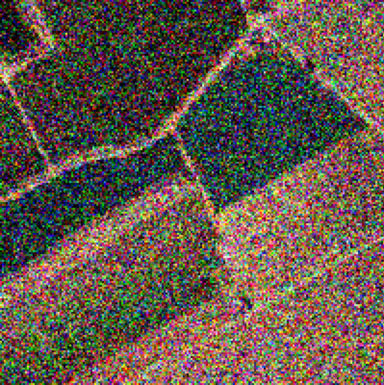}}
        \medskip
    \end{minipage}
    \hfill
    \begin{minipage}[b]{0.19\linewidth}
        \centering
        \centerline{\includegraphics[width=3.3cm]{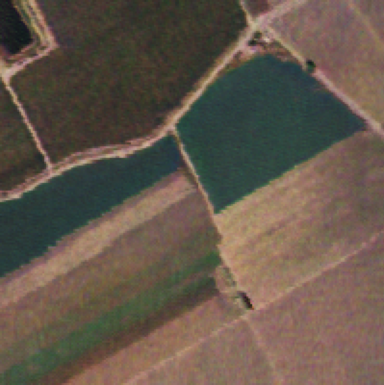}}
        \medskip
    \end{minipage}
    \hfill
    \begin{minipage}[b]{0.19\linewidth}
        \centering
        \centerline{\includegraphics[width=3.3cm]{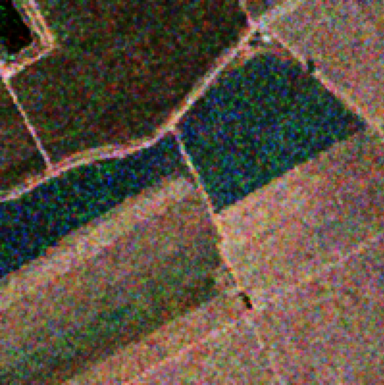}}
        \medskip
    \end{minipage}
    \hfill
    \begin{minipage}[b]{0.19\linewidth}
        \centering
        \centerline{\includegraphics[width=3.3cm]{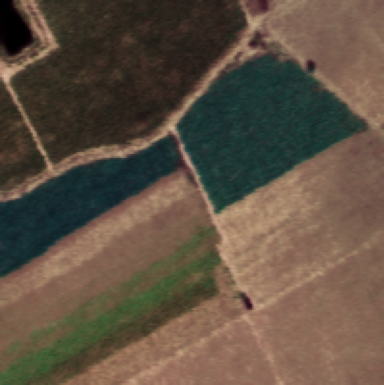}}
        \medskip
    \end{minipage}
    \\
    \begin{minipage}[b]{0.19\linewidth}
        \centering
        \centerline{\includegraphics[width=3.3cm]{media/5-1.png}}
        \medskip
    \end{minipage}
    \hfill
    \begin{minipage}[b]{0.19\linewidth}
        \centering
        \centerline{\includegraphics[width=3.3cm]{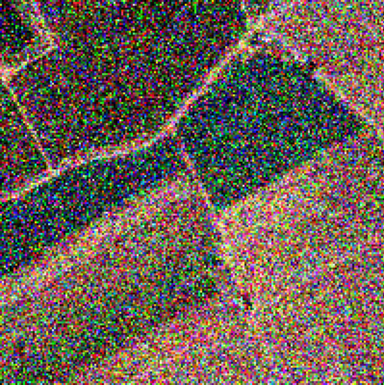}}
        \medskip
    \end{minipage}
    \hfill
    \begin{minipage}[b]{0.19\linewidth}
        \centering
        \centerline{\includegraphics[width=3.3cm]{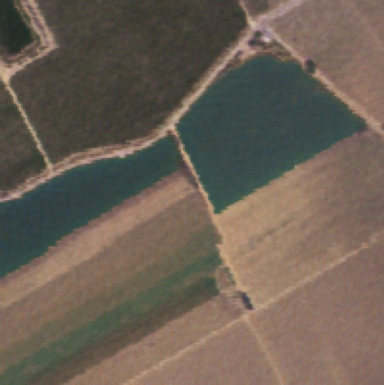}}
        \medskip
    \end{minipage}
    \hfill
    \begin{minipage}[b]{0.19\linewidth}
        \centering
        \centerline{\includegraphics[width=3.3cm]{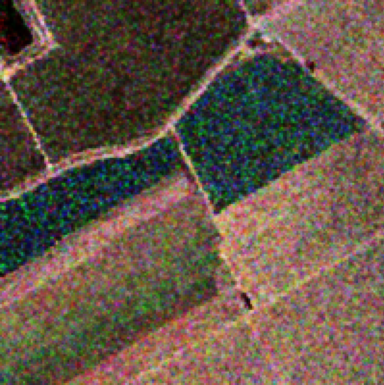}}
        \medskip
    \end{minipage}
    \hfill
    \begin{minipage}[b]{0.19\linewidth}
        \centering
        \centerline{\includegraphics[width=3.3cm]{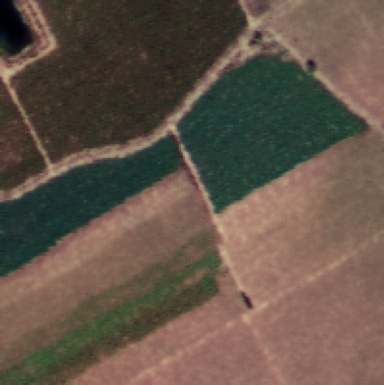}}
        \medskip
    \end{minipage}
    \\
    \begin{minipage}[b]{0.19\linewidth}
        \centering
        \centerline{\includegraphics[width=3.3cm]{media/5-1.png}}
        \centerline{(a) Original}
    \end{minipage}
    \hfill
    \begin{minipage}[b]{0.19\linewidth}
        \centering
        \centerline{\includegraphics[width=3.3cm]{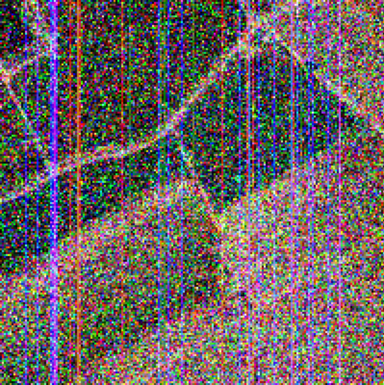}}
        \centerline{(b) Noisy}
    \end{minipage}
    \hfill
    \begin{minipage}[b]{0.19\linewidth}
        \centering
        \centerline{\includegraphics[width=3.3cm]{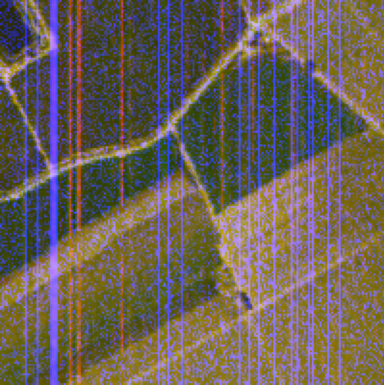}}
        \centerline{(c) SURE-DHIP \cite{nguyen2021}}
    \end{minipage}
    \hfill
    \begin{minipage}[b]{0.19\linewidth}
        \centering
        \centerline{\includegraphics[width=3.3cm]{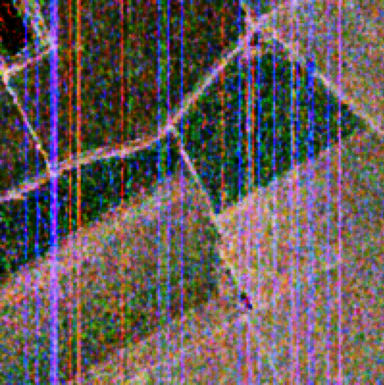}}
        \centerline{(d) HLF-DHIP \cite{niresi2022}}
    \end{minipage}
    \hfill
    \begin{minipage}[b]{0.19\linewidth}
        \centering
        \centerline{\includegraphics[width=3.3cm]{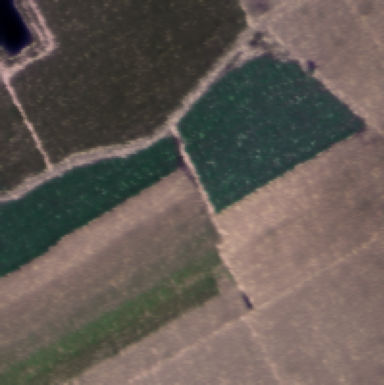}}
        \centerline{(e) Proposed}
    \end{minipage}
    \caption{Original Salinas HSI segment, noisy version, and denoised versions obtained by SURE-DHIP \cite{nguyen2021}, HLF-DHIP \cite{niresi2022} and the proposed method. The noise content in each row is Gaussian, Gaussian + sparse, and Gaussian + sparse + stripes for rows 1, 2, and 3 respectively.}
    \label{fig:5}
\end{figure*}

Figures \ref{fig:4} and \ref{fig:5} provide a qualitative comparison of the denoised outputs at the final training iteration for each of the two datasets. False-color visualizations are generated using spectral bands 56, 26, and 16 (Washington DC Mall) and 29, 19, 9 (Salinas). For the DC Mall segment, the proposed method produces visibly cleaner reconstructions for each noise scenario, with reduced residual noise and fewer artifacts compared to both SURE-DHIP and HLF-DHIP, corroborating the quantitative results. For the Salinas segment, the proposed method is again very competitive, especially in scenarios with stripe noise, which it successfully eliminates, in contrast to SURE-DHIP and HLF-DHIP. In scecnarios containing only Gaussian or Gaussian + Sparse noise, we observe that SURE-DHIP is competitive to the proposed method, while HLF-DHIP shows poorer results. This is attributed to the low-rankness of the Salinas image, which somewhat disfavors the use of the pure HLF. However, even in this unfavorable scenario, our method remains robust and achieves excellent denoising results.

Finally, Tables I-IV and V-VIII summarize the MPSNR and MSSIM values at the final iteration for Gaussian, Gaussian + sparse, Gaussian + stripe, and combined noise scenarios respectively, for each of the two datasets. In the Washington DC Mall HSI, the proposed method consistently achieves the best performance across all SNR levels and noise types, while competing methods either suffer from reduced robustness or overfitting. In the case of the Salinas image, the proposed method directly challenges and often outperforms SURE-DHIP, despite the favorable performance of the latter in comparison to HLF-DHIP, as mentioned above. These results confirm that the proposed loss formulation effectively balances robustness and sensitivity control, enabling stable denoising performance under realistic and challenging noise conditions. It should be noted that all three methods exhibit comparable running times, while the proposed approach allows safe early termination due to its inherent robustness to overfitting.

\section{Conclusions}

This paper addressed overfitting in deep hyperspectral image prior (DHIP)-based denoising by combining robust data fidelity with explicit sensitivity regularization under joint optimization of the network parameters and the input. The proposed Smooth $\ell_1$-divergence loss prevents noise memorization, yields stable convergence, and consistently outperforms existing DIP-based methods across Gaussian, sparse, and stripe noise without relying on early stopping. Future work will investigate theoretical justification of the observed behavior and explore alternative network architectures for sensitivity-controlled unsupervised reconstruction.

\begin{table}[ht]
    \caption{MPSNR and MSSIM results of the three methods on \textbf{Washington DC Mall} HSI for Gaussian noise.}
    \footnotesize
    \centering
    \renewcommand\arraystretch{1.2}
    \begin{tabular}{c|c c c c}
        \toprule
        \textbf{SNR (dB)} & \textbf{Noisy} & \textbf{SURE-DIP}\cite{nguyen2021} & \textbf{HLF-DIP}\cite{niresi2022} & \textbf{Proposed} \\
        \midrule
        \multirow{2}{*}{\textbf{0}}  
            & 13.15 & 24.06 & 21.12 & \textbf{27.77} \\
            & 0.196 & 0.812 & 0.584 & \textbf{0.845} \\
        \midrule
        \multirow{2}{*}{\textbf{5}}  
            & 17.47 & 30.03 & 27.56 & \textbf{32.13} \\
            & 0.381 & 0.918 & 0.841 & \textbf{0.934} \\
        \midrule
        \multirow{2}{*}{\textbf{10}} 
            & 21.96 & \textbf{36.01} & 33.54 & 35.70 \\
            & 0.598 & \textbf{0.971} & 0.951 & 0.967 \\
        \bottomrule
    \end{tabular}
    \label{tab:1}
\end{table}
\begin{table}[!ht]
    \caption{MPSNR and MSSIM results of the three methods on \textbf{Washington DC Mall} HSI for Gaussian + sparse noise.}
    \footnotesize
    \centering
    \renewcommand\arraystretch{1.2}
    \begin{tabular}{c|c c c c}
        \toprule
        \textbf{SNR (dB)} & \textbf{Noisy} & \textbf{SURE-DIP}\cite{nguyen2021} & \textbf{HLF-DIP}\cite{niresi2022} & \textbf{Proposed} \\
        \midrule
        \multirow{2}{*}{\textbf{0}}  
            & 11.02 & 19.85 & 20.06 & \textbf{25.89} \\
            & 0.143 & 0.657 & 0.535 & \textbf{0.796} \\
        \midrule
        \multirow{2}{*}{\textbf{5}}  
            & 13.05 & 23.49 & 26.20 & \textbf{30.71} \\
            & 0.232 & 0.805 & 0.802 & \textbf{0.920} \\
        \midrule
        \multirow{2}{*}{\textbf{10}} 
            & 14.18 & 24.17 & 32.56 & \textbf{34.98} \\
            & 0.305 & 0.765 & 0.942 & \textbf{0.965} \\
        \bottomrule
    \end{tabular}
    \label{tab:2}
\end{table}
\begin{table}[!ht]
    \caption{MPSNR and MSSIM results of the three methods on \textbf{Washington DC Mall} HSI for Gaussian + stripe noise.}
    \footnotesize
    \centering
    \renewcommand\arraystretch{1.2}
    \begin{tabular}{c|c c c c}
        \toprule
        \textbf{SNR (dB)} & \textbf{Noisy} & \textbf{SURE-DIP}\cite{nguyen2021} & \textbf{HLF-DIP}\cite{niresi2022} & \textbf{Proposed} \\
        \midrule
        \multirow{2}{*}{\textbf{0}}
            & 12.87 & 22.48 & 20.99 & \textbf{27.51} \\
            & 0.191 & 0.744 & 0.590 & \textbf{0.844} \\
        \midrule
        \multirow{2}{*}{\textbf{5}}  
            & 16.83 & 27.32 & 26.55 & \textbf{31.51} \\
            & 0.366 & 0.853 & 0.804 & \textbf{0.930} \\
        \midrule
        \multirow{2}{*}{\textbf{10}} 
            & 20.72 & 31.51 & 31.39 & \textbf{33.02} \\
            & 0.569 & 0.906 & 0.902 & \textbf{0.937} \\
        \bottomrule
    \end{tabular}
    \label{tab:3}
\end{table}
\begin{table}[!ht]
    \caption{MPSNR and MSSIM results of the three methods on \textbf{Washington DC Mall} HSI for Gaussian + sparse + stripe noise.}
    \footnotesize
    \centering
    \renewcommand\arraystretch{1.2}
    \begin{tabular}{c|c c c c}
        \toprule
        \textbf{SNR (dB)} & \textbf{Noisy} & \textbf{SURE-DIP}\cite{nguyen2021} & \textbf{HLF-DIP}\cite{niresi2022} & \textbf{Proposed} \\
        \midrule
        \multirow{2}{*}{\textbf{0}}
            & 10.87 & 19.38 & 20.09 & \textbf{25.00} \\
            & 0.140 & 0.671 & 0.550 & \textbf{0.785} \\
        \midrule
        \multirow{2}{*}{\textbf{5}}  
            & 12.79 & 21.94 & 25.91 & \textbf{30.06} \\
            & 0.225 & 0.732 & 0.788 & \textbf{0.914} \\
        \midrule
        \multirow{2}{*}{\textbf{10}} 
            & 13.86 & 22.97 & 30.57 & \textbf{34.28} \\
            & 0.295 & 0.732 & 0.890 & \textbf{0.962} \\
        \bottomrule
    \end{tabular}
    \label{tab:4}
\end{table}
\begin{table}[ht]
    \caption{MPSNR and MSSIM results of the three methods on \textbf{Salinas} HSI for Gaussian noise.}
    \footnotesize
    \centering
    \renewcommand\arraystretch{1.2}
    \begin{tabular}{c|c c c c}
        \toprule
        \textbf{SNR (dB)} & \textbf{Noisy} & \textbf{SURE-DIP}\cite{nguyen2021} & \textbf{HLF-DIP}\cite{niresi2022} & \textbf{Proposed} \\
        \midrule
        \multirow{2}{*}{\textbf{0}}  
            & 8.89 & 21.05 & 15.20 & \textbf{23.85} \\
            & 0.042 & \textbf{0.611} & 0.146 & 0.606 \\
        \midrule
        \multirow{2}{*}{\textbf{5}}  
            & 11.88 & 26.98 & 21.11 & \textbf{28.95} \\
            & 0.087 & \textbf{0.748} & 0.320 & 0.743 \\
        \midrule
        \multirow{2}{*}{\textbf{10}} 
            & 16.03 & \textbf{32.97} & 27.05 & 32.57 \\
            & 0.171 & \textbf{0.882} & 0.590 & 0.838 \\
        \bottomrule
    \end{tabular}
    \label{tab:5}
\end{table}
\begin{table}[!ht]
    \caption{MPSNR and MSSIM results of the three methods on \textbf{Salinas} HSI for Gaussian + sparse noise.}
    \footnotesize
    \centering
    \renewcommand\arraystretch{1.2}
    \begin{tabular}{c|c c c c}
        \toprule
        \textbf{SNR (dB)} & \textbf{Noisy} & \textbf{SURE-DIP}\cite{nguyen2021} & \textbf{HLF-DIP}\cite{niresi2022} & \textbf{Proposed} \\
        \midrule
        \multirow{2}{*}{\textbf{0}}  
            & 8.58 & 20.00 & 14.51 & \textbf{22.71} \\
            & 0.039 & 0.575 & 0.137 & \textbf{0.628} \\
        \midrule
        \multirow{2}{*}{\textbf{5}}  
            & 11.10 & 24.92 & 20.31 & \textbf{27.31} \\
            & 0.075 & \textbf{0.757} & 0.296 & 0.719 \\
        \midrule
        \multirow{2}{*}{\textbf{10}} 
            & 14.06 & 27.84 & 25.98 & \textbf{31.02} \\
            & 0.134 & \textbf{0.824} & 0.548 & 0.801 \\
        \bottomrule
    \end{tabular}
    \label{tab:6}
\end{table}
\begin{table}[!ht]
    \caption{MPSNR and MSSIM results of the three methods on \textbf{Salinas} HSI for Gaussian + stripe noise.}
    \footnotesize
    \centering
    \renewcommand\arraystretch{1.2}
    \begin{tabular}{c|c c c c}
        \toprule
        \textbf{SNR (dB)} & \textbf{Noisy} & \textbf{SURE-DIP}\cite{nguyen2021} & \textbf{HLF-DIP}\cite{niresi2022} & \textbf{Proposed} \\
        \midrule
        \multirow{2}{*}{\textbf{0}}
            & 8.89 & 20.88 & 15.25 & \textbf{22.86} \\
            & 0.042 & \textbf{0.600} & 0.149 & 0.598 \\
        \midrule
        \multirow{2}{*}{\textbf{5}}  
            & 11.83 & 26.73 & 20.87 & \textbf{28.51} \\
            & 0.086 & \textbf{0.773} & 0.316 & 0.732 \\
        \midrule
        \multirow{2}{*}{\textbf{10}} 
            & 15.85 & 30.92 & 26.59 & \textbf{31.51} \\
            & 0.168 & \textbf{0.817} & 0.583 & 0.785 \\
        \bottomrule
    \end{tabular}
    \label{tab:7}
\end{table}
\begin{table}[!t]
    \caption{MPSNR and MSSIM results of the three methods on \textbf{Salinas} HSI for Gaussian + sparse + stripe noise.}
    \footnotesize
    \centering
    \renewcommand\arraystretch{1.2}
    \begin{tabular}{c|c c c c}
        \toprule
        \textbf{SNR (dB)} & \textbf{Noisy} & \textbf{SURE-DIP}\cite{nguyen2021} & \textbf{HLF-DIP}\cite{niresi2022} & \textbf{Proposed} \\
        \midrule
        \multirow{2}{*}{\textbf{0}}
            & 8.59 & 18.91 & 14.48 & \textbf{22.86} \\
            & 0.039 & 0.424 & 0.138 & \textbf{0.630} \\
        \midrule
        \multirow{2}{*}{\textbf{5}}  
            & 11.08 & 24.46 & 20.12 & \textbf{26.76} \\
            & 0.075 & \textbf{0.739} & 0.297 & 0.715 \\
        \midrule
        \multirow{2}{*}{\textbf{10}} 
            & 13.95 & 26.85 & 25.43 & \textbf{30.69} \\
            & 0.132 & 0.789 & 0.536 & \textbf{0.801} \\
        \bottomrule
    \end{tabular}
    \label{tab:8}
\end{table}

\bibliographystyle{IEEEtran}
\bibliography{refs}

\end{document}